\pdfoutput=1
\documentclass[11pt]{article}

\usepackage[]{ACL2023}
\usepackage{graphicx}

\usepackage{times}
\usepackage{natbib}
\usepackage{latexsym}
\usepackage{amsmath}
\usepackage{multirow}
\usepackage{tabularx} % Required for full width table
\usepackage{booktabs} % For professional looking tables

\usepackage[T1]{fontenc}
\usepackage[utf8]{inputenc}

\usepackage{microtype}

\usepackage{inconsolata}

\title{Control Token with Dense Passage Retrieval}

\author{
  Juhwan Lee \\
  Diqest \\
  \texttt{9521ljh@gmail.com} \\\And
  Jisu Kim \\
  Diqest \\
  \texttt{jisukim8873@gmail.com} \\}
  
\begin{document}
\maketitle
\begin{abstract}
This study addresses the hallucination problem in large language models (LLMs). We adopted Retrieval-Augmented Generation (RAG)~\citep{lewis2020retrieval}, a technique that involves embedding relevant information in the prompt to obtain accurate answers. However, RAG also faced inherent issues in retrieving correct information. To address this, we employed the Dense Passage Retrieval (DPR)~\citep{karpukhin2020dense} model for fetching domain-specific documents related to user queries. Despite this, the DPR model still lacked accuracy in document retrieval. We enhanced the DPR model by incorporating control tokens, achieving significantly superior performance over the standard DPR model, with a 13\% improvement in Top-1 accuracy and a 4\% improvement in Top-20 accuracy.
\end{abstract}

\section{Introduction}

Large language models (LLMs) have recently gained significant attention. Unlike traditional models, LLMs can perform a wide variety of tasks and exhibit performance comparable to pre-trained language models (PLMs), making them highly esteemed for their versatility and excellence~\citep{brown2020language}. However, LLMs still face various challenges and limitations, such as hallucination—generating incorrect information—and biases that can lead to social issues. Many AI companies are striving to address these issues, with hallucination being a particularly critical concern due to the probabilistic nature of deep learning models. Deep learning models can generate plausible yet incorrect information~\citep{sun2024trustllm}.

A promising approach to mitigate this issue is RAG, which involves retrieving relevant information and embedding it in the prompt to guide the LLM's responses. This method underscores the renewed importance of search engines, which must fetch accurate and relevant information for LLMs to produce correct answers. Numerous studies have demonstrated the effectiveness of RAG in reducing hallucination~\citep{gao2023retrieval}. Consequently, enhancing system components like search engines is crucial for improving LLM service quality. This study focuses on developing a search engine tailored for LLMs, employing a deep learning-based search engine that can be specialized for specific domains.

\section{Related Work}

Various techniques exist in the search engine domain. Simple methods include keyword-based similarity searches like TF-IDF (Term Frequency-Inverse Document Frequency) and BM25. These methods assess lexical similarity based on word frequency and inverse frequency, offering quick and algorithmically sound search capabilities by reflecting the importance of specific words. However, from an LLM perspective, the diverse expression of user queries makes it increasingly challenging to search for similar documents based solely on specific words. 

Recently, embedding models that convert sentences into vectors and search for semantically similar documents have garnered attention. Examples include word embedding methodologies and the SentenceTransformer model~\citep{thakur-2020-AugSBERT}, which are trained on large datasets to effectively extract semantically similar sentences. However, these methods may underperform in very specific domains due to insufficient domain-specific data training. To address this, domain-specific search engine models like Dense Passage Retrieval (DPR) have emerged. The DPR model is trained to assign high similarity scores to relevant information and low scores to irrelevant information, allowing for effective fine-tuning to cater to specific user data requirements. The DPR model has shown superior performance over BM25, making it a suitable choice for our research aimed at developing a domain-specific search engine by enhancing the DPR model. 

\begin{figure*}[ht]
    \centering
    \includegraphics[width=0.8\textwidth]{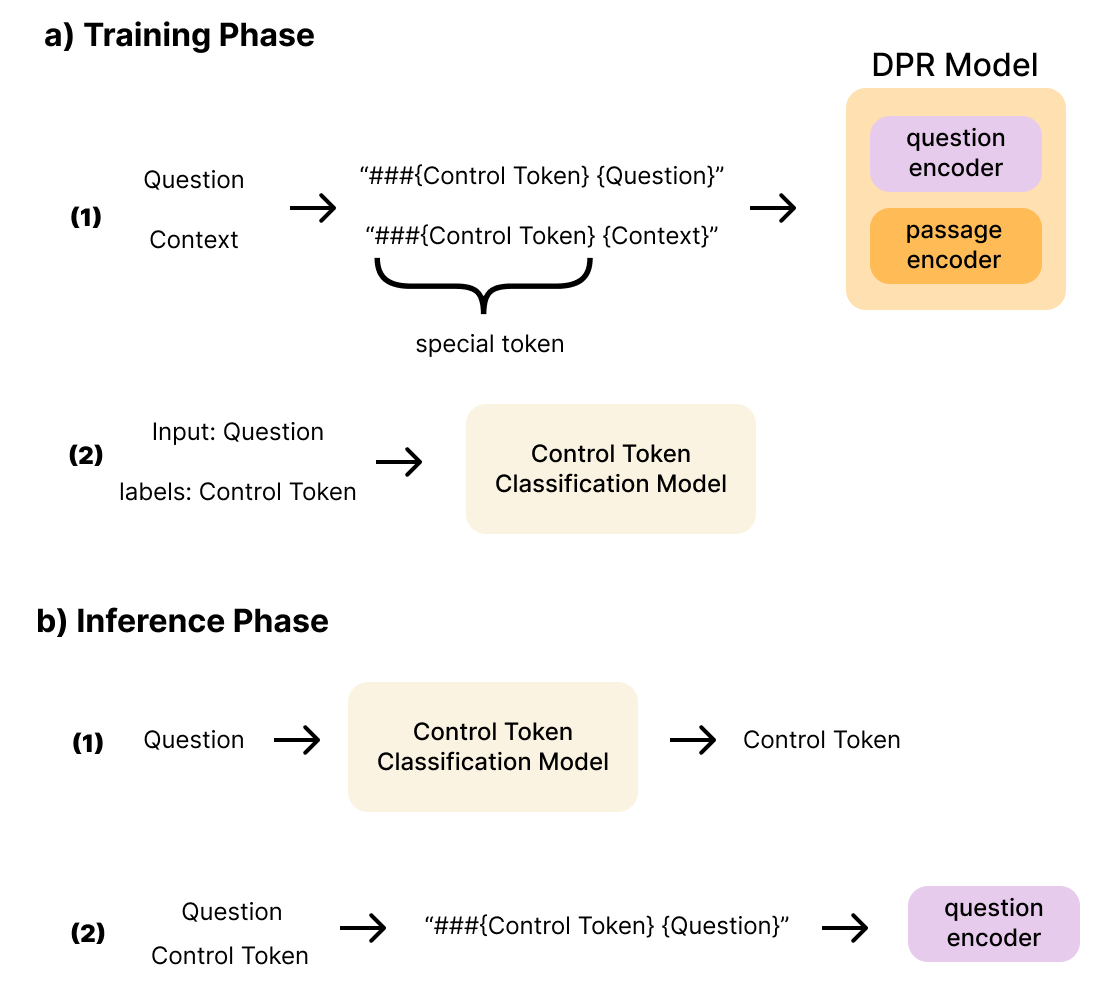}
    \caption{
    \textbf{a) Training phase:} (1) Combine CT with query and context: CT is combined with the query and context as a special token for training the DPR model. (2) Train CT classification model: A classification model is trained to extract the correct CT from the query.
    \textbf{b) Inference phase:} (1) CT classification: Use the trained CT classification model to extract the CT for the query. (2) Combine CT with query: Combine the extracted CT with the query and use the DPR model's question encoder to derive vectors.
    }
    \label{fig:figure1}
\end{figure*}

\section{Method}
The core of a chatbot's search engine is understanding user intent. Accurate comprehension of user intent is essential for narrowing the search scope and providing highly accurate search results. However, the existing DPR model's limitation is its lack of reflection on user intent, relying solely on text similarity. Thus, we devised a novel methodology combining user intent with the DPR model. This approach, illustrated in Figure 1, defines "intent" as a specific class for each document. This idea stemmed from the multilingual translation model, NLLB-200~\citep{costa2022no}, which uses language tokens in inputs for translation. We adapted this concept by including "intent" related tokens in the input, termed Control Tokens (CT). 

Our research objective is to fine-tune the DPR model to enhance search accuracy by incorporating CT in the input. Specifically, CT represents the user's query intent related to a particular document. For example, when a user queries about a specific domain, adding the domain-specific CT to the input allows the model to provide more accurate search results. We trained the DPR model with CT-included inputs to develop a system capable of effectively retrieving documents related to user intent, significantly improving search accuracy. Figure 1 shows the entire process flow. We refer to this combined model of CT and DPR as cDPR.

% Your other sections here...

\subsection*{Data}
We selected an MRC dataset containing questions, context, answers, and categories suitable for our experiment. As a Korean company, we chose AIHUB's administrative dataset, specializing in Korean language data. This dataset includes a diverse range of complex documents, totaling 63,930 entries, with 11 categories based on administrative documents, including science and technology, public administration, land management, and so on. The dataset can be accessed at AIHUB's website\footnote{\url{https://aihub.or.kr/aihubdata/data/view.do?currMenu=115&topMenu=100&aihubDataSe=data&dataSetSn=569}}.

\subsection*{Context Chunking} 
We chunked the dataset's context into sentences, then expanded the context around the index of the answer sentence to 512 tokens.

\subsection*{Control Token}
We defined document classes as Control Tokens (CT). To create special tokens, we prefixed classes with "\#\#\#" and applied them to the tokenizer and model. We added CTs to the training data without altering the DPR model's architecture. If the classification threshold was not met, we treated the CT as "[unk]" token.

\subsection*{Control Token Classification}
We developed a classifier that learns the categories of documents required to answer a question from the same AIHUB MRC dataset. This classifier categorizes the document class based on the question. We fine-tuned a pre-trained model, FacebookAI/xlm-roberta-base~\citep{DBLP:journals/corr/abs-1911-02116}, using our dataset to achieve this. During the inference phase, it provides the appropriate control token for the query.

\subsection*{Dense Passage Retrieval}
We retained the original DPR model structure and trained it with CT-included data, as shown in Figure 1, by fine-tuning the encoder model using Klue/roberta-base~\citep{park2021klue}.

\section{Results}
Our evaluation checked if the retrieved context contained the correct answer using the MRC dataset. The CT classification model achieved 74\% accuracy on the dataset. We tested various thresholds for CT classification, observing that the enhanced CT-attached training data improved Top-20 accuracy to 93.79\%, a 4\% increase over the standard DPR model, with a 13\% improvement in Top-1 accuracy. Higher CT classification model performance correlated with better retrieval accuracy, highlighting CT's critical role in search engines.

The comparison between the standard DPR model and our proposed cDPR model across different top-N retrieval accuracies is presented in Table \ref{table:comparison}. It is evident from the results that cDPR outperforms the DPR base model significantly, particularly at Top-1 retrieval, demonstrating the effectiveness of CT-attached training data in improving model performance.

\begin{table}[]
\centering
\begin{tabular}{|l|l|l|}
\hline
      & DPR base & cDPR(ours) \\ \hline
Top1  & 51.1\%   & 64.4\%     \\ \hline
Top5  & 77.8\%   & 85.7\%     \\ \hline
Top10 & 84.9\%   & 90.4\%     \\ \hline
Top15 & 87.8\%   & 92.5\%     \\ \hline
Top20 & 89.6\%   & 93.8\%     \\ \hline
\end{tabular}
\caption{Comparison of Retrieval Performance between DPR and cDPR Models}
\label{table:comparison}
\end{table}

Furthermore, we explored the impact of various classification thresholds on the cDPR model's performance, as shown in Table \ref{table:classification_thresholds}. The results indicate that increasing the classification threshold generally enhances retrieval performance across all top-N levels. Specifically, a threshold of 0.9 resulted in the highest Top-1 accuracy of 64.4\%, reinforcing the importance of an effective classification threshold in improving overall model performance.

\begin{table}[]
\centering
\begin{tabular}{|l|c|c|c|c|}
\hline
\multirow{2}{*}{cDPR} & \multicolumn{4}{c|}{classification threshold} \\ \cline{2-5} 
                      & \(\geq 0\) & \(\geq 0.5\) & \(\geq 0.7\) & \(\geq 0.9\) \\ \hline
Top1                  & 61.7\%     & 62.5\%       & 63.4\%       & 64.4\%       \\ \hline
Top5                  & 83.6\%     & 84.2\%       & 85.0\%       & 85.7\%       \\ \hline
Top10                 & 88.3\%     & 88.8\%       & 89.6\%       & 90.4\%       \\ \hline
Top15                 & 90.9\%     & 91.4\%       & 91.9\%       & 92.5\%       \\ \hline
Top20                 & 92.3\%     & 92.6\%       & 93.3\%       & 93.8\%       \\ \hline
\end{tabular}
\caption{Classification Performance at Different Thresholds}
\label{table:classification_thresholds}
\end{table}

\section{Discussion}
This study demonstrated the importance and role of Control Tokens (CT) in search models. Although we used CT as a special token representing a question category, incorporating multiple CT types reflecting keywords, intent, and context could further enhance performance. This approach is particularly useful in conversational AI systems like chatbots, contributing to developing advanced search engines that accommodate diverse user intents, thereby enabling accurate and efficient information retrieval across various domains.

\bibliography{anthology,custom}
\bibliographystyle{acl_natbib}

\end{document}